\documentclass[11pt,a4paper]{article}
\usepackage[hyperref]{emnlp-ijcnlp-2019}
\usepackage{times}
\usepackage{latexsym}
\usepackage{graphicx}
\usepackage{amsmath}
\usepackage{url}
\usepackage{tabularx}
\usepackage[shortlabels]{enumitem}

\aclfinalcopy 

\makeatletter
\newcommand{\printfnsymbol}[1]{%
  \textsuperscript{\@fnsymbol{#1}}%
}
\makeatother

\title{Identifying Supporting Facts for Multi-hop Question Answering with Document Graph Networks}
\author{Mokanarangan Thayaparan\thanks{\ \ : \ \texttt{equal contribution}} ,  Marco Valentino\printfnsymbol{1}, Viktor Schlegel\printfnsymbol{1}, Andr\'e Freitas \\
  Department of Computer Science\\ University of Manchester\\
  {\tt \{thayaparan.mokanarangan,marco.valentino,viktor.schlegel,andre.freitas\}}\\{\tt @manchester.ac.uk}
  }

\date{}

\begin{document}
\maketitle
\begin{abstract}
Recent advances in reading comprehension have resulted in  models that surpass human performance when the answer is contained in a single, continuous passage of text. However, complex Question Answering (QA) typically requires multi-hop reasoning -- i.e. the integration of supporting facts from different sources, to infer the correct answer.

This paper proposes Document Graph Network (DGN), a message passing architecture for the identification of supporting facts over a graph-structured representation of text.

The evaluation on HotpotQA shows that DGN obtains competitive results when compared to a reading comprehension baseline operating on raw text, confirming the relevance of structured representations for supporting multi-hop reasoning.

\end{abstract}

\section{Introduction}

Question Answering (QA) is the task of inferring the answer for a natural language question in a given knowledge source. Acknowledged as a suitable task for benchmarking natural language understanding, QA is gradually evolving from mere retrieval task to a well-established tool for testing complex forms of reasoning. Recent advances in deep learning have sparked interest in a specific type of QA emphasising Machine Comprehension (MC) aspects, where background knowledge is entirely expressed in form of unstructured text.  
\begin{figure}[t]
\centering
\includegraphics[width=\columnwidth]{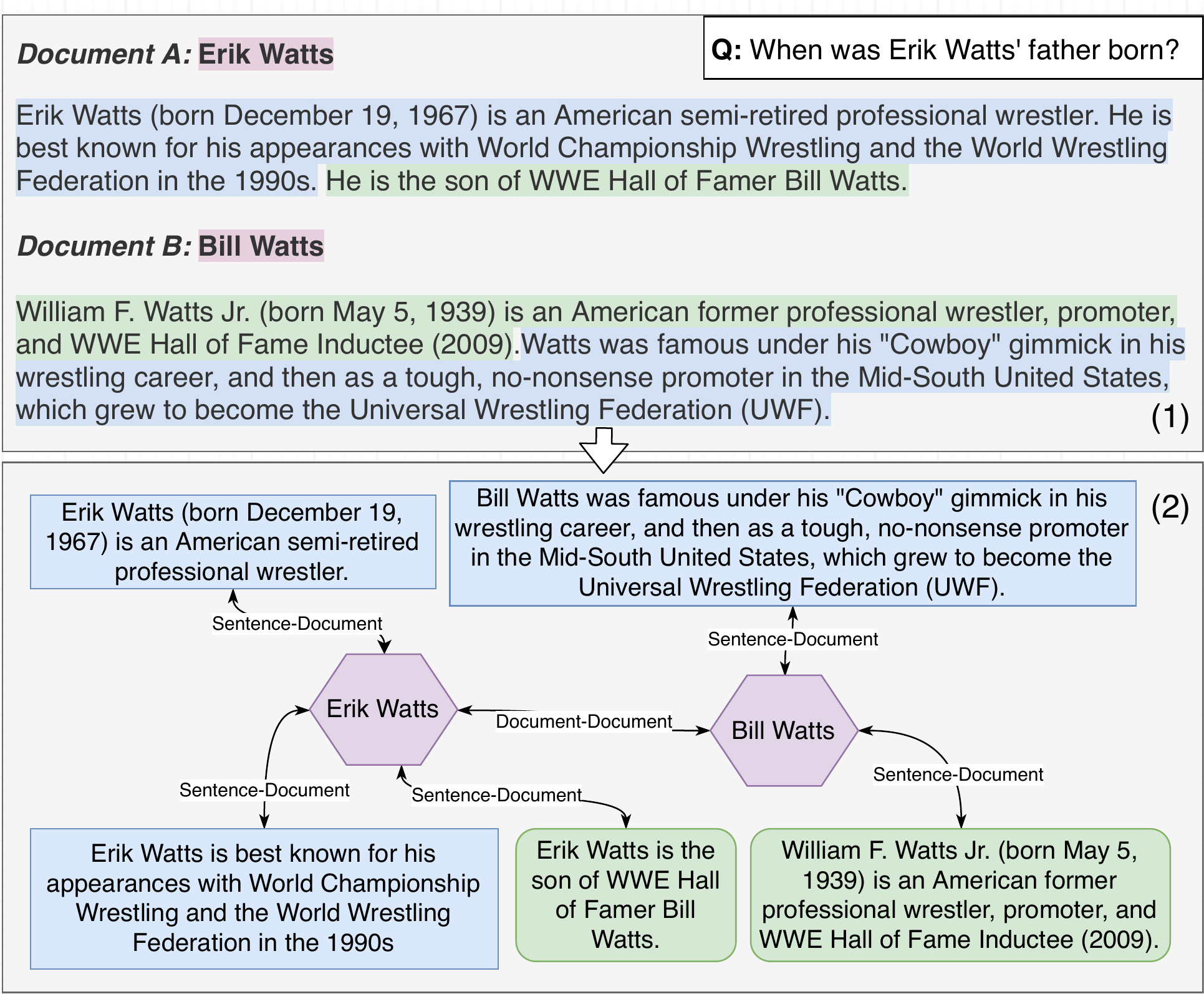}
\caption{Is structure important for complex, multi-hop Question Answering (QA) over unstructured text passages? To answer this question we explore the task of identifying supporting facts \textcolor{green}{(rounded rectangles)} by transforming a corpus of documents (1) into an undirected graph (2) connecting sentence nodes \textcolor{blue}{(rectangles)} and document nodes \textcolor{purple}{(hexagons)}. }
\label{fig:document_graph}
\end{figure}

State-of-the-art techniques for MC typically retrieve the answer from a continuous passage of text by adopting a combination of character and word-level models with various forms of attention mechanisms \cite{seo2016bidirectional, yu2018qanet}. By employing unsupervised pre-training on large corpora \cite{devlin2018bert}, these models are capable of outperforming humans in reading comprehension tasks where the context is represented by a single paragraph \cite{rajpurkar2018know}.

However, when it comes to answering complex questions on large document collections, it is unlikely that a single passage can provide sufficient evidence to support the answer. Complex QA typically requires multi-hop reasoning, i.e. the ability of combining multiple information fragments from different sources.

Moreover, recent studies have raised concerns on inference capabilities, generalisation and interpretability of current MC models \cite{wiese2017neural,dhingra2017quasar,kaushik2018much}, leading to the creation of novel datasets that propose multi-hop reading comprehension as a benchmark for evaluating complex reasoning and explainability \cite{yang2018hotpotqa}.

Consider the example in Figure \ref{fig:document_graph}. In order to answer the question \emph{``When was Erik Watts' father born?"} a QA system has to retrieve and combine supporting facts stored in different documents:
\begin{enumerate}
    \item Document A: \emph{``Erik Watts is the son of WWE Hall of Famer Bill Watts''};
    \item Document B: \emph{``William F. Watts Jr. (born May 5, 1939) is an American former professional wrestler, promoter, and WWE Hall of Fame Inductee (2009)''}.
\end{enumerate}
The explicit selection of supporting facts has a dual role in a multi-hop QA pipeline:
\begin{enumerate}[(a)]
    \item It allows the system to consider all and only those facts that are relevant to answer a specific question;
    \item It provides an explicit trace of the reasoning process, which can be presented as justification for the answer.
\end{enumerate}

This paper explores the task of identifying supporting facts for multi-hop QA over large collections of documents where several passages act as distractors for the MC model.
In this setting, we hypothesise that graph-structured representations play a key role in reducing complexity and improving the ability to retrieve meaningful evidence for the answer. 

As shown in Figure \ref{fig:document_graph}.1, identifying supporting facts in unstructured text is challenging as it requires capturing long-term dependencies to exclude irrelevant passages. On the other hand (Figure \ref{fig:document_graph}.2), a graph-structured representation connecting related documents simplifies the integration of relevant facts by making them mutually reachable in few hops. We put this observation in practice by transforming a text corpus in a global representation that links documents and sentences by means of mutual references. 

In order to identify supporting facts on undirected graphs, we investigate the use of message passing architectures with relational inductive bias \cite{battaglia2018relational}. We present the Document Graph Network (DGN), a specific type of Gated Graph Neural Network (GGNN) \cite{li2015gated} trained to identify supporting facts in the aforementioned structured representation.

We evaluate DGN on HotpotQA \cite{yang2018hotpotqa}, a  recently proposed dataset for assessing MC performance on supporting facts identification. 
The experiments show that DGN is able to obtain improvements in F1 score when compared to a MC baseline that adopts a sequential reading strategy. 
The obtained results confirm the value of pursuing research towards the definition of novel MC architectures, which are able to incorporate structure as an integral part of their learning and inference processes.

\section{Document Graph Network}
\begin{figure*}
\centering
\includegraphics[width=1.04\textwidth]{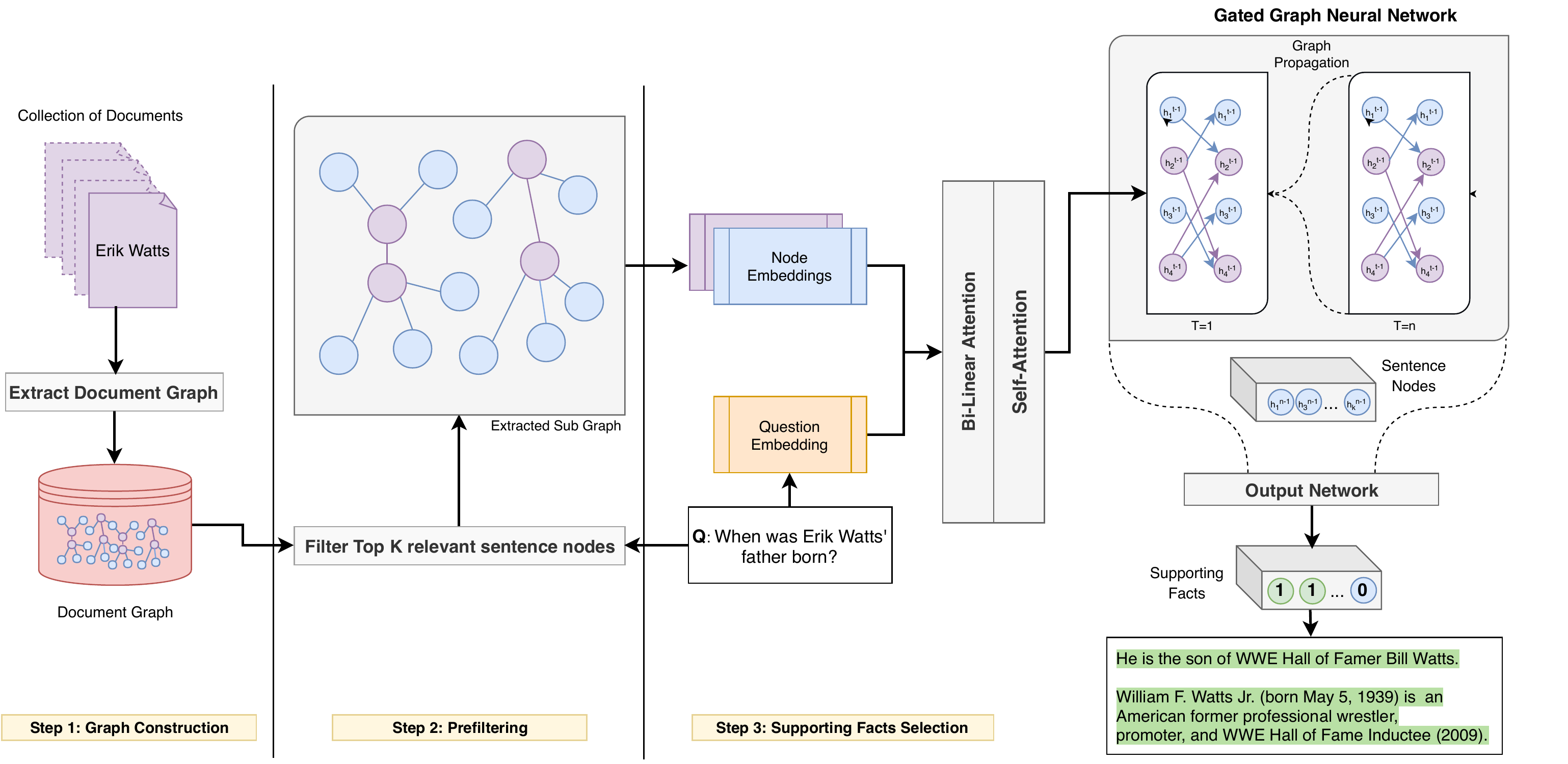}
\caption{Overview of the approach for the identification of supporting facts in a multi-hop QA pipeline. Step 1 is applied offline for extracting a graph-structured representation from large corpora (Sec. \ref{sec:representation}). In Step 2, we employ a filtering algorithm (Sec. \ref{sec:filtering}) to retrieve a sub-graph containing the top $k$ relevant sentences nodes. The final step (Step 3) adopts the DGN model for message passing and binary classification of the supporting facts (Sec. \ref{sec:sf_selection}).}
\label{fig:approach}
\end{figure*}

The following section presents the Document Graph Network (DGN), a message passing architecture designed to identify supporting facts for multi-hop QA on graph-structured representations of documents.

Here, we discuss in details the construction of the underlying graph, the DGN model, and a pre-filtering step implemented to alleviate the impact of large graphs on model complexity.

\subsection{Graph-structured Representation}
\label{sec:representation}
Given an arbitrary corpus of documents $\mathcal{D} = \{D_1, D_2, \dots , D_n\}$,  we aim to build an 
undirected document graph $DG$ as structured representation of $\mathcal{D}$ (Figure \ref{fig:document_graph}). 

The advantage of using graph-structured representations lies in reducing the inference steps necessary to combine two or more supporting facts. Therefore, we want to extract a representation that increases the probability of connecting relevant sentences with short paths in the graph. We observe that multi-hop questions usually require reasoning on two concepts/entities that are described in different, but interlinked documents. We put in practice this observation by connecting two documents if they contain mentions to the same entities.

The Document Graph (DG) contains nodes of two types. We represent each document $D_i$ in $\mathcal{D}$ as a document node $d_i$ and each of its sentences $S_{jD_i} $ as a sentence node $s_{jD_i}$. We then add an edge of type $e_{sentence-document}$ that links them. This edge type represents the fact that a specific sentence belongs to a specific document. We apply coreference resolution to solve implicit entity mentions within the documents. Subsequently, we add an edge of type $e_{document-document}$ between two document nodes $d_1, d_2$, if the entities described in $D_1$ are referenced in $D_2$ or vice-versa.

Given a question $q$, we use $DG$ (instead of $\mathcal{D}$) as input for the DGN model. the representation does not include edges between sentences since we observed increasing complexity in the model without gaining substantial benefits in terms of performance.

\subsection{Architectural Overview}
\label{sec:architecture}

Figure~\ref{fig:approach} highlights the main components of the DGN architecture.

From the target corpus, we automatically extract a Document Graph $DG$ encoding the background knowledge expressed in a corpus of documents (Step 1). This data and its graphical structure is permanently stored into a database, ready to be loaded when it is required. The first step is performed offline, allowing the integration of new knowledge regardless of the runtime pipeline implemented to address the task.

In order to speed up the computation and alleviate current drawbacks of Gated Graph Neural Networks \cite{li2015gated}, the question answering pipeline is augmented with a prefiltering step (Step 2). The adopted algorithm (Sec \ref{sec:filtering}), based on a relevance score, is aimed at reducing the number of nodes involved in the computation.
Current limitations of Gated Graph Neural Networks, in fact, are mainly connected with the size of the input graph used for learning and prediction. Performance in terms of computational efficiency and learning degrades proportionally to the number of nodes and edge types in the input graph. 
In order to reduce the negative impact of large graphs, we adopt the prefiltering step to prune $DG$, and retrieve a set of sentence nodes $\mathcal{S} = \{S_1, S_2, \dots, S_k\}$ expected to contain supporting facts for a question $q$. 

The subsequent step (Step 3) is aimed at selecting supporting facts for $q$. For this task we employ the Document Graph Network (DGN) on the subset of $DG$ induced by $\mathcal S$ (section \ref{sec:sf_selection}). Specifically, we apply the aforementioned architecture to learn a distributed representation of each node in the graph via message passing. This representation is then used by an Output Network (ON) to perform binary classification on the sentence nodes in $\mathcal{S}$ and select a set of supporting facts $SF = \{sf_1, sf_2, \dots, sf_m\}$ with $ SF \subseteq \mathcal S$. In the experiments we perform supervised learning on the training set provided by HotpotQA \cite{yang2018hotpotqa} to correctly predict the elements belonging to $SF$.

\subsection{Prefiltering Step}
\label{sec:filtering}
 Given a question $q$ and a set of documents $D = \{D_1, D_2, \dots , D_n\}$ as context, the aim of the prefiltering step is to retrieve a subset of the context containing the $k$ most relevant sentences to $q$. 
 
 In order to achieve this goal, we adopt a ranking based approach similar to the one illustrated in \cite{narasimhan2018out}. Specifically, we consider all the sentences occurring in the documents and compute the similarity between each word in a sentence and each word in the question $q$. We adopt pre-trained GloVe vectors \cite{pennington2014glove} to obtain the distributed representation of each word. 
Subsequently, we produce the relevance score of each sentence by calculating the mean among the $m$ highest similarity values. The final subset is obtained by selecting the sentences with the top $k$ relevance scores.

An empirical analysis suggested that $m = 5$ gives the best results on the development set. We evaluated this approach by computing the recall of retrieving the top $k$ supporting facts for $k = \{20,25,30\}$, obtaining values greater than $90\%$ for $k = 25$ and $k = 30$. Since the average number of candidate sentences for each question in the corpus is 50.89, the described algorithm allows us to discard $60.7\%$ ($k=20$), $50.87\%$ ($k=25$) and $41.05\%$ ($k=30$) of irrelevant context. 

\subsection{Identifying Supporting Facts}
The Document Graph Network (DGN)  is employed for the identification of supporting facts. The DGN model is  based on a standard Gated Graph Neural Network architecture (GGNN) \cite{li2015gated} where the inner representation of the nodes is customised to carry out this specific task. We apply DGN on the sub-graph retrieved by the filtering module.

In alignment with prior research in the field we encode Question($Q$), Nodes($N$) and Graph($G$) as follows:

\begin{enumerate}
    \item \textbf{Question Representation}: The question is stripped of punctuation and stop words and tokenised to obtain $W$ words. These words are subsequently converted into a tensor $Q \in R^{|W| \times D}$ using pre-trained GloVe vectors \cite{pennington2014glove} of dimension $D$.
    \item \textbf{Node Representation}: Similar to the question representation, each node is also converted to $ V \in R^{|W| \times D}$ using entities for document nodes and sentences words for sentence nodes.
    \item \textbf{Graph Representation}: Each document graph $DG$ is represented by an adjacency matrix $A \in R^{|V| \times 2|E||V|}$ where $V$ and $E$ denote the vertices and edge types respectively.
\end{enumerate}

Each node ($v_i$) is conditioned on the question ($q_i$) using Bi-Linear Attention~\cite{kim2018bilinear}. The attention weights $\alpha_{i}$ of each word $w$ in the nodes are determined by a learned function $f_{BAN}$ as shown in Equation 2. Here $f_{BAN}$ computes the attention scores between two matrices using a bilinear attention function. This function has a matrix of weights $W$ and a bias vector $b$ used to calculate the similarity between the two matrices as $VWQ^T + b$:
\begin{align}
\begin{split}
     e_{iw} &= f_{BAN}(v_{iw}, q_i)\\
\end{split}\\
\begin{split}\label{biattn_score}
    \alpha_{iw} &= \frac{exp(e_{iw})}{\sum_{k=1}^{|W|} exp(e_{ik})}\\
\end{split}
\end{align}

Following the calculation of the attention scores, the question conditioned vectors are determined as follows:
\begin{equation}
    \hat{v}_i = \phi(\{v_i\},\{\alpha_i\})
\end{equation}

Here, $\phi$ is a learned function that combines the attention scores of each word by employing a non-linear transformation.

After conditioning the nodes representation on the question, we employ a Self-Attention Model function $f_{SAN}$~\cite{vaswani2017attention} to calculate the weight of each vector $\delta_i$. Here, the learned function $f_{SAN}$ is responsible for computing the weights of each vector in a node. The rationale behind this operation is to condense the matrices to a vector suitable for a Gated Graph Neural Network architecture while retaining the most discriminative semantic information.

\begin{align}
\begin{split}
   r_{iw}  &= f_{SAN}(\hat{v}_{iw})\\
\end{split}\\
\begin{split}
    \delta_{iw} &= \frac{exp(r_{iw})}{\sum_{k=1}^{|W|} exp(r_{ik})}\\
\end{split}
\end{align}

After computing the self-attention score, we calculate the initial annotation vectors for the GGNN as follows:
\begin{equation}
    x_{v} = \sigma(\{\hat{v}_i\},\{\delta_i\})
\end{equation}
where $\sigma$ is a function that returns a single vector by multiplying the corresponding attention scores and summing them up.
\label{sec:sf_selection}
The basic recurrent unit of a GGNN can be formalised as follows:

\begin{align}
h_{v}^{(1)} &= [x_{v}^{T}, 0]^{T}\\
a_{v}^{(t)} &= A_{v:}^{T}[h_{1}^{(t-1)T} ...h_{|V|}^{(t-1)T}]^{T} + b\\
z_{v}^{t} &= \sigma(W^{z}a_{v}^{(t)} + U^{z}h_{v}^{(t-1)})\\
r_{v}^{t} &= \sigma(W^{r}a_{v}^{(t)} + U^{r}h_{v}^{(t-1)})\\
\widetilde{h_{v}^{t}} &= tanh(Wa^{(t)}_{v} + U(r^{t}_v \odot h_{v}^{(t-1)}))\\
h^{(t)}_{v} &= (1 - z_{v}^{t}) \odot h_{v}^{(t-1)} + z_{v}^{t} \odot \widetilde{h_{v}^{(t)}}
\end{align}

We perform $T$ time steps of propagation and retrieve the distributed nodes representation by using the final hidden state. The computed representation of each node implicitly captures the semantic information of its neighbours at a distance up to $T$ hops. In the experiments, we found it sufficient to set $T = 3$. 

The graph is heterogeneous with nodes representing questions, sentences and documents. As the supporting facts identification task requires sentence classification, we retain the final hidden state of the sentence nodes while discarding the others. We use the sentence representations as input to a feed forward neural network called Output Network. We perform binary classification of each sentence to predict whether it is a supporting fact or not:

\begin{equation} 
o_v = g(h_{v}^{(T)}, x_v)
\end{equation}

\section{Evaluation}
The experiments are motivated by the guiding research question of the paper: \emph{Does structure play a role in identifying supporting facts for multi-hop Question Answering?} We further break down the question in the following research hypotheses: 
\begin{itemize}
    \item \textbf{RH1:} Existing machine comprehension models benefit from reducing the context to a small number of sentences necessary to answer a question.
    \item \textbf{RH2:} Models operating on a graph-structured representation perform better, supporting the identification of relevant facts when compared to a baseline that uses a sequential strategy.
\end{itemize}
We seek to provide evidence for those claims by conducting the following experiments:
\begin{itemize}
    \item \textbf{Experiment 1:} investigate how a representative state-of-the-art MC model performs on different passages with varying coherency and length.
    \item \textbf{Experiment 2:} evaluate the capability of the proposed approach to identify supporting facts in a question answering scenario where the relevant facts are distributed across multiple documents.
\end{itemize}

Specific tests are performed to identify contributing features and compare the overall performance of the approach with a sequential baseline reported in the literature.

\paragraph{HotpotQA}
We ran the experiments over the recently proposed HotpotQA dataset \cite{yang2018hotpotqa}, which requires MC models to find supporting passages in a large set of documents, and perform multi-hop reasoning to arrive at the correct answer. HotpotQA provides 105,547~first~paragraphs extracted from Wikipedia articles, and corresponding question-answer pairs created by human annotators. Questions are designed to only be answerable by combining information from two articles and require to bridge documents via a concept or entity mentioned in both articles. A subset of questions require a comparison of similar concepts concerning their common or differing properties. Furthermore, the dataset provides labels for supporting sentences, making it possible to perform quantitative analysis on the retrieval of supporting facts. 

In all of the reported experiments, if not stated otherwise, training is performed on the HotpotQA \emph{training set} while the evaluation is performed on the \emph{development set} in the \emph{distractor setting}. In order to address this setting, a system has to retrieve the answer and the supporting facts for a given question by reasoning over a set of ten documents. Only two of the supplied documents are guaranteed to contain the information that is sufficient and necessary to answer the question. The remaining eight documents are similar documents retrieved by an information retrieval model (hence the name \emph{distractor}).

\subsection{State-of-the-Art Machine Comprehension Performance}
This experiment is designed to investigate the capabilities of single passage MC models to retrieve the correct answer when provided with a context of varying size and coherency. For this analysis we adopt BERT \cite{devlin2018bert}, a neural transformer architecture \cite{vaswani2017attention} constituting the  state-of-the-art latent representation for various NLP tasks. 

The publicly available model is pre-trained in an unsupervised manner on a large text corpus with the objective of language modelling and next sentence prediction. Fine-tuning this model to specific NLP tasks has shown to achieve state-of-the-art-results for many NLP tasks, among others question answering and machine reading comprehension  \cite{devlin2018bert}. To that end, we fine-tune the model on the training split of HotpotQA and evaluate it on the evaluation split. Before training, we manually remove all the questions that cannot be answered by retrieving a continuous passage in the supporting facts (e.g. we exclude comparison questions that typically require yes/no type of answers). 

We evaluate the performance of BERT with supporting facts only, and then progressively enrich the context by a rising number of sentences retrieved by the filtering algorithm (Sec. \ref{sec:filtering}). The results of this experiment are reported in Table~\ref{tab:experiment_1}.

Note that these results can not be interpreted as a resilient comparison baseline as \emph{(1)} we don't optimise the set of hyper-parameters associated with the model training and \emph{(2)} we ensure the existence of supporting facts in the evaluation, since we are interested in the intrinsic performance of BERT in answer retrieval.

Unsurprisingly, the best results are achieved when the context provided to BERT is composed of supporting facts only. Conversely, the performance of the model gradually deteriorates when distracting information is added to the context.

These results reinforce our assumption that a module capable of identifying the correct set of supporting facts represents a fundamental component in a multi-hop QA pipeline. Moreover, this component may be complementary to downstream machine comprehension models, constituting a valid support to improve overall performances in answer retrieval.

\begin{table}
\caption{F1 and exact match (EM) score of BERT when evaluated on answer retrieval over contexts of varying size.}
\label{tab:experiment_1}
\centering
\resizebox{\columnwidth}{!}{\begin{tabular}{lcccccc}
\hline
\textbf{Sentences}  & \textbf{SF only} & \textbf{+5} & \textbf{+10} & \textbf{+20} & \textbf{+25} & \textbf{+30} \\
\hline
 \textbf{F1} & \textbf{0.75} & 0.60 & 0.52 & 0.44 & 0.42 & 0.42\\
\hline
 \textbf{EM} & \textbf{0.60} & 0.47 & 0.40 & 0.33 & 0.32 & 0.31\\
 \hline
\end{tabular}}
\end{table}

\subsection{Supporting Facts Identification}
\begin{table}
\caption{Supporting facts identification: Harmonic mean (\textbf{F1}), \textbf{P}recision and \textbf{R}ecall}
\label{tab:sf}
\centering
\resizebox{\columnwidth}{!}{\begin{tabular}{lccc}
\hline
\textbf{Model} & \textbf{F1} & \textbf{P} & \textbf{R} \\
\hline
Baseline \cite{yang2018hotpotqa} & 66.66 & - & - \\
\hline
Baseline Replication (Answer + SF) &  65.28 & \textbf{63.28} & 67.43\\
\hline
Baseline Replication (SF only) & 46.44 & 48.80 & 44.31 \\
\hline
DGN (best) & \textbf{68.02} & 61.51 & \textbf{76.07} \\
\hline
DGN (no edge types) & 45.84 & - & - \\
\hline
\end{tabular}}
\end{table}

We compare the DGN model on the task of identifying supporting facts against the neural baseline reported in \cite{yang2018hotpotqa}. In order to suit the task, the baseline architecture extends the state-of-the-art answer passage retrieval model \cite{seo2016bidirectional} by an additional recurrent layer that classifies whether a sentence is a supporting fact or not. The model is trained jointly and under strong supervision on the objectives of retrieving both answer and supporting facts. We replicate the experiment on our infrastructure in order to obtain more detailed measures, such as precision an recall. The results of the evaluation are reported in Table~\ref{tab:sf}. 

The experiments show that the DGN model outperforms the baseline in terms of F1 score ($\approx$2\% improvement compared to the results reported in the paper, $\approx$3\% improvement compared to our replication), and recall ($\approx$14\% improvement over our replication). 
However, the baseline implementation has a higher precision. 
We attribute that to the fact that the baseline optimises for both answer extraction and supporting facts retrieval. 

In general we observe that recall is higher than precision throughout the experiments. 
Compared to the DGN model, the baseline is less penalised when the retrieved answer still matches the expected answer, even if retrieved from an unrelated sentence spuriously. in the absence of the answer selection optimisation criterion, the DGN model is only penalised if it fails to predict the correct supporting facts. This forces the model to prioritise recall over precision during training. 
Adding a weight to the loss calculation as an additional hyperparameter can balance the precision and recall metric.

We don't evaluate DGN on the task of answer retrieval, since the proposed architecture focuses on the classification of the relevant supporting facts. The task of jointly retrieving answer and supporting facts is left for future work.

\subsection{Analysis}

\begin{table}
\caption{\textbf{P}recision, \textbf{R}ecall and \textbf{F1} score of the DGN model with different values of $k$ assigned to the prefiltering algorithm. The best values are highlighted.}
\label{tab:filtering_recall}
\centering
\resizebox{\columnwidth}{!}{\begin{tabular}{lrrrr}
\hline
  & \textbf{k = 20} & \textbf{k = 25} & \textbf{k = 30} & \textbf{full}\\
\hline
 \textbf{Prefiltering (R)} & 84.72 & 90.23 & 94.29 & 100.00\\
 \hline
 \textbf{DGN (F1)} & 63.40 & 67.38 & \textbf{68.02} & 66.00 \\
 \textbf{DGN (P)} & 57.83 & \textbf{61.92} & 61.51 & 53.37 \\
 \textbf{DGN (R)} & 70.17 & 73.90 & 76.07 & \textbf{86.47} \\
\hline
\end{tabular}}
\end{table}
In order to understand the interaction of the key contributing parts of the architecture, we analyse the behaviour of the full pipeline in different settings. Specifically, we measure the DGN performance when trained and evaluated on the output of the filtering step. During the training, we ensure the existence of the supporting facts in the input graph of the DGN model. We then evaluate it on the development set by performing prediction on the subset retrieved by the filtering algorithm. The results reported in Table \ref{tab:filtering_recall} take into account the combined performance of the full pipeline with different hyperparameters assigned to the prefiltering algorithm.

Firstly, we observe the increase of recall with the increasing number of retrieved sentences. This fact is unsurprising and it is in line with the higher recall score of the filtering module. More sentences means broader coverage, and thus higher recall even before executing DGN prediction. 

Secondly, across the experiments, we observe that $k = 30$ is the best number of sentences for the model to learn from. This is confirmed by the best precision and overall F1 score obtained when training and predicting on the top 30 sentences. Moreover, we observed that the application of the filtering algorithm sensibly speeds up the training, decreasing at the same time the amount of memory required to store matrices and weights of the graph network. The application of a light filtering is then justified both in terms of performance and computational complexity.

Regarding the baseline model, we aim to analyse the impact of multi-task learning, where the model is jointly trained to retrieve supporting facts and the final answer. We observe a significant drop in performance ($\approx$20\% F1 score) when we optimise the baseline only for supporting facts identification (see \emph{Baseline Replication} in Table~\ref{tab:sf}). This observation is perfectly in line with the literature. \cite{hashimoto2016joint} report improvements on low level tasks when jointly optimised with higher level tasks in a hierarchical learning setting. Regarding multi-hop QA, the identification of supporting facts directly depends on the answer being predicted correctly and vice-versa. A plausible future work may be to understand whether DGN can benefit from a similar multi-task learning setup.

Finally, we investigate the role of the semantic information expressed explicitly in the Document Graph. To that end, we train the DGN model using the same configuration of the best performing model without edge type information. This results in a notable drop of F1 score (see Table~\ref{tab:sf}) reinforcing the evidence that explicit semantic information encoded in relational form contributes towards the performance of the model. A promising future direction will be to investigate whether different types of semantic representation benefit the performance of the model and to what extent.

\section{Related Work}
State-of-the-art approaches for Open-Domain Question Answering over large collections of documents employ a combination of character-level models, self-attention \cite{wang2017gated}, and bi-attention \cite{seo2016bidirectional} to operate over unstructured paragraphs without exploiting any structured text representation. Despite these methods have demonstrated impressive results reaching in some cases super-human performances \cite{seo2016bidirectional,chen2017reading,yu2018qanet}, recent studies have raised important concerns related to generalisation \cite{wiese2017neural,dhingra2017quasar} complex reasoning \cite{welbl2018constructing} and explainability \cite{yang2018hotpotqa}. Specifically, the lack of structured representation makes it hard for current Machine Comprehension models to find meaningful patterns in large corpora, generalise beyond the training domain and justify the answer. 

Research efforts towards the creation of message-passing architectures with relational inductive bias \cite{battaglia2018relational} have enabled machine learning algorithms to incorporate graphical structures in their training process. These models, trained over explicit entities and relations, have the potential to boost generalisation, interpretability and abstract reasoning capabilities. A variety of Graph Neural Network architectures have already demonstrated remarkable results in a large set of applications ranging from Computer Vision, Physical Systems and Protein-Protein Interaction \cite{zhou2018graph}. 

Our research is in line with recent trends in Question Answering prone to explore message-passing architectures over graph-structured representation of documents to enhance performance and overcome challenges involved in dealing with unstructured text. \cite{sun2018open} fuse text corpus with manually-curated knowledge bases to create heterogeneous graphs of KB facts and text sentences. Their model, GRAFT-Net, built upon Graph Convolutional Networks \cite{schlichtkrull2018modeling}, is used to propagate information between heterogeneous nodes in the graph and perform binary classification on entity nodes to select the answer. Differently from the proposed approach, the latter work focuses on links between whole paragraphs and external entities in a Knowledge Base. Moreover, GRAFT-Net is designed for single-hop Question Answering, assuming that the question is always about a single entity.

The proposed approach is similar to \cite{de2018question} and \cite{song2018exploring}, where the aim is to answer complex questions that require the integration of multiple text passages.
However, our research is focused on the identification of supporting facts instead of answer retrieval.

Another line of research focuses on narrowing down the context for later Machine Comprehension models by selecting relevant passages as supporting facts. Work in that direction includes \cite{watanabe2017question} which present a neural information retrieval system to retrieve a sufficiently small paragraph and \cite{geva2018learning} which employ a Deep Q-Network (DQN) to solve the task by learning to navigate over an intra-document tree. A similar approach is chosen by  \cite{clark2017simple}. However, instead of operating on document structure, they adopt a sampling technique to make the model more robust towards multi-paragraph documents.
These approaches are not directly comparable to our work since they focus either on single paragraphs or intra-document (local) structure.

Strongly related to our work is \cite{yang2018hotpotqa} which presents HotpotQA, a novel dataset for multi-hop QA. The authors highlight the importance of identifying supporting facts for improving reasoning and explainability of current systems. We compare the proposed architecture with the baseline described in their paper. The model is based on a state-of-the-art MC model \cite{seo2016bidirectional} that adopts a sequential reading strategy to identifying supporting facts from large collections of documents.

\section{Conclusion}
In this paper, we investigated the role played by interlinked sentence representation for complex, multi-hop question answering under the focus of supporting facts identification, i.e. retrieving the minimum set of facts required to answer a given question. We emphasise that this problem is worth pursuing, showing that the performance of state-of-the-art models substantially deteriorates as the size of the accompanying context increases.

We present Document Graph Network (DGN), a novel approach for selecting supporting facts in a multi-hop QA pipeline. The model operates over explicit relational knowledge, connecting documents and sentences extracted from large text corpora. We adopt a pre-filtering step to limit the number of nodes and train a customised Graph Gated Neural Network directly on the extracted representation.

We train and evaluate the DGN model on a newly proposed dataset for complex, multi-hop question answering over unstructured text. The evaluation shows that DGN outperforms a baseline adopting a sequential reading strategy. Additionally, we show that when trained to retrieve just supporting facts, the performance of the baseline degrades by $\approx$20\%.

Perhaps most importantly, we highlight a way to combine structured and distributional sentence representation models and propose further research lines in that direction. As future work, we aim to investigate the role and impact of different structured sentence representation models within the inference process, linking it with the Open Information Extraction \cite{cetto2018graphene, niklaus-etal-2018-survey} and sentence simplification \cite{niklaus-etal-2019-transforming, niklaus2017sentence} literature.

We believe that further research can be dedicated to inject richer structured knowledge in the model, allowing for fine-grained message passing and improved representation learning. Another important line of research will focus on the implementation of advanced mechanisms and techniques to scale the approach to massive text corpora such as the whole Wikipedia.

\subsection*{Acknowledgements}
The authors would like to express their gratitude towards members of the AI Systems lab at the University of Manchester for many fruitful and intense discussions.
\bibliography{emnlp-ijcnlp-2019}

\begin{thebibliography}{30}
\expandafter\ifx\csname natexlab\endcsname\relax\def\natexlab#1{#1}\fi

\bibitem[{Battaglia et~al.(2018)Battaglia, Hamrick, Bapst, Sanchez-Gonzalez,
  Zambaldi, Malinowski, Tacchetti, Raposo, Santoro, Faulkner
  et~al.}]{battaglia2018relational}
Peter~W Battaglia, Jessica~B Hamrick, Victor Bapst, Alvaro Sanchez-Gonzalez,
  Vinicius Zambaldi, Mateusz Malinowski, Andrea Tacchetti, David Raposo, Adam
  Santoro, Ryan Faulkner, et~al. 2018.
\newblock Relational inductive biases, deep learning, and graph networks.
\newblock \emph{arXiv preprint arXiv:1806.01261}.

\bibitem[{Cetto et~al.(2018)Cetto, Niklaus, Freitas, and
  Handschuh}]{cetto2018graphene}
Matthias Cetto, Christina Niklaus, Andr{\'e} Freitas, and Siegfried Handschuh.
  2018.
\newblock Graphene: Semantically-linked propositions in open information
  extraction.
\newblock \emph{Prooceedings of COLING}.

\bibitem[{Chen et~al.(2017)Chen, Fisch, Weston, and Bordes}]{chen2017reading}
Danqi Chen, Adam Fisch, Jason Weston, and Antoine Bordes. 2017.
\newblock Reading wikipedia to answer open-domain questions.
\newblock \emph{arXiv preprint arXiv:1704.00051}.

\bibitem[{Clark and Gardner(2017)}]{clark2017simple}
Christopher Clark and Matt Gardner. 2017.
\newblock Simple and effective multi-paragraph reading comprehension.
\newblock \emph{arXiv preprint arXiv:1710.10723}.

\bibitem[{De~Cao et~al.(2018)De~Cao, Aziz, and Titov}]{de2018question}
Nicola De~Cao, Wilker Aziz, and Ivan Titov. 2018.
\newblock Question answering by reasoning across documents with graph
  convolutional networks.
\newblock \emph{arXiv preprint arXiv:1808.09920}.

\bibitem[{Devlin et~al.(2018)Devlin, Chang, Lee, and
  Toutanova}]{devlin2018bert}
Jacob Devlin, Ming-Wei Chang, Kenton Lee, and Kristina Toutanova. 2018.
\newblock Bert: Pre-training of deep bidirectional transformers for language
  understanding.
\newblock \emph{arXiv preprint arXiv:1810.04805}.

\bibitem[{Dhingra et~al.(2017)Dhingra, Mazaitis, and Cohen}]{dhingra2017quasar}
Bhuwan Dhingra, Kathryn Mazaitis, and William~W Cohen. 2017.
\newblock Quasar: Datasets for question answering by search and reading.
\newblock \emph{arXiv preprint arXiv:1707.03904}.

\bibitem[{Geva and Berant(2018)}]{geva2018learning}
Mor Geva and Jonathan Berant. 2018.
\newblock Learning to search in long documents using document structure.
\newblock \emph{arXiv preprint arXiv:1806.03529}.

\bibitem[{Hashimoto et~al.(2016)Hashimoto, Xiong, Tsuruoka, and
  Socher}]{hashimoto2016joint}
Kazuma Hashimoto, Caiming Xiong, Yoshimasa Tsuruoka, and Richard Socher. 2016.
\newblock A joint many-task model: Growing a neural network for multiple nlp
  tasks.
\newblock \emph{arXiv preprint arXiv:1611.01587}.

\bibitem[{Kaushik and Lipton(2018)}]{kaushik2018much}
Divyansh Kaushik and Zachary~C Lipton. 2018.
\newblock How much reading does reading comprehension require? a critical
  investigation of popular benchmarks.
\newblock \emph{arXiv preprint arXiv:1808.04926}.

\bibitem[{Kim et~al.(2018)Kim, Jun, and Zhang}]{kim2018bilinear}
Jin-Hwa Kim, Jaehyun Jun, and Byoung-Tak Zhang. 2018.
\newblock Bilinear attention networks.
\newblock In \emph{Advances in Neural Information Processing Systems}, pages
  1571--1581.

\bibitem[{Li et~al.(2015)Li, Tarlow, Brockschmidt, and Zemel}]{li2015gated}
Yujia Li, Daniel Tarlow, Marc Brockschmidt, and Richard Zemel. 2015.
\newblock Gated graph sequence neural networks.
\newblock \emph{arXiv preprint arXiv:1511.05493}.

\bibitem[{Narasimhan et~al.(2018)Narasimhan, Lazebnik, and
  Schwing}]{narasimhan2018out}
Medhini Narasimhan, Svetlana Lazebnik, and Alexander Schwing. 2018.
\newblock Out of the box: Reasoning with graph convolution nets for factual
  visual question answering.
\newblock In \emph{Advances in Neural Information Processing Systems}, pages
  2659--2670.

\bibitem[{Niklaus et~al.(2017)Niklaus, Bermeitinger, Handschuh, and
  Freitas}]{niklaus2017sentence}
Christina Niklaus, Bernhard Bermeitinger, Siegfried Handschuh, and André
  Freitas. 2017.
\newblock \href {http://arxiv.org/abs/1703.09013} {A sentence simplification
  system for improving relation extraction}.

\bibitem[{Niklaus et~al.(2018)Niklaus, Cetto, Freitas, and
  Handschuh}]{niklaus-etal-2018-survey}
Christina Niklaus, Matthias Cetto, Andr{\'e} Freitas, and Siegfried Handschuh.
  2018.
\newblock A survey on open information extraction.
\newblock In \emph{Proceedings of the 27th International Conference on
  Computational Linguistics}, pages 3866--3878, Santa Fe, New Mexico, USA.
  Association for Computational Linguistics.

\bibitem[{Niklaus et~al.(2019)Niklaus, Cetto, Freitas, and
  Handschuh}]{niklaus-etal-2019-transforming}
Christina Niklaus, Matthias Cetto, Andr{\'e} Freitas, and Siegfried Handschuh.
  2019.
\newblock \href {https://doi.org/10.18653/v1/P19-1333} {Transforming complex
  sentences into a semantic hierarchy}.
\newblock In \emph{Proceedings of the 57th Annual Meeting of the Association
  for Computational Linguistics}, pages 3415--3427, Florence, Italy.
  Association for Computational Linguistics.

\bibitem[{Pennington et~al.(2014)Pennington, Socher, and
  Manning}]{pennington2014glove}
Jeffrey Pennington, Richard Socher, and Christopher Manning. 2014.
\newblock Glove: Global vectors for word representation.
\newblock In \emph{Proceedings of the 2014 conference on empirical methods in
  natural language processing (EMNLP)}, pages 1532--1543.

\bibitem[{Rajpurkar et~al.(2018)Rajpurkar, Jia, and Liang}]{rajpurkar2018know}
Pranav Rajpurkar, Robin Jia, and Percy Liang. 2018.
\newblock Know what you don't know: Unanswerable questions for squad.
\newblock \emph{arXiv preprint arXiv:1806.03822}.

\bibitem[{Schlichtkrull et~al.(2018)Schlichtkrull, Kipf, Bloem, van~den Berg,
  Titov, and Welling}]{schlichtkrull2018modeling}
Michael Schlichtkrull, Thomas~N Kipf, Peter Bloem, Rianne van~den Berg, Ivan
  Titov, and Max Welling. 2018.
\newblock Modeling relational data with graph convolutional networks.
\newblock In \emph{European Semantic Web Conference}, pages 593--607. Springer.

\bibitem[{Seo et~al.(2016)Seo, Kembhavi, Farhadi, and
  Hajishirzi}]{seo2016bidirectional}
Minjoon Seo, Aniruddha Kembhavi, Ali Farhadi, and Hannaneh Hajishirzi. 2016.
\newblock Bidirectional attention flow for machine comprehension.
\newblock \emph{arXiv preprint arXiv:1611.01603}.

\bibitem[{Song et~al.(2018)Song, Wang, Yu, Zhang, Florian, and
  Gildea}]{song2018exploring}
Linfeng Song, Zhiguo Wang, Mo~Yu, Yue Zhang, Radu Florian, and Daniel Gildea.
  2018.
\newblock Exploring graph-structured passage representation for multi-hop
  reading comprehension with graph neural networks.
\newblock \emph{arXiv preprint arXiv:1809.02040}.

\bibitem[{Sun et~al.(2018)Sun, Dhingra, Zaheer, Mazaitis, Salakhutdinov, and
  Cohen}]{sun2018open}
Haitian Sun, Bhuwan Dhingra, Manzil Zaheer, Kathryn Mazaitis, Ruslan
  Salakhutdinov, and William Cohen. 2018.
\newblock Open domain question answering using early fusion of knowledge bases
  and text.
\newblock In \emph{Proceedings of the 2018 Conference on Empirical Methods in
  Natural Language Processing}, pages 4231--4242.

\bibitem[{Vaswani et~al.(2017)Vaswani, Shazeer, Parmar, Uszkoreit, Jones,
  Gomez, Kaiser, and Polosukhin}]{vaswani2017attention}
Ashish Vaswani, Noam Shazeer, Niki Parmar, Jakob Uszkoreit, Llion Jones,
  Aidan~N Gomez, {\L}ukasz Kaiser, and Illia Polosukhin. 2017.
\newblock Attention is all you need.
\newblock In \emph{Advances in Neural Information Processing Systems}, pages
  5998--6008.

\bibitem[{Wang et~al.(2017)Wang, Yang, Wei, Chang, and Zhou}]{wang2017gated}
Wenhui Wang, Nan Yang, Furu Wei, Baobao Chang, and Ming Zhou. 2017.
\newblock Gated self-matching networks for reading comprehension and question
  answering.
\newblock In \emph{Proceedings of the 55th Annual Meeting of the Association
  for Computational Linguistics (Volume 1: Long Papers)}, volume~1, pages
  189--198.

\bibitem[{Watanabe et~al.(2017)Watanabe, Dhingra, and
  Salakhutdinov}]{watanabe2017question}
Yusuke Watanabe, Bhuwan Dhingra, and Ruslan Salakhutdinov. 2017.
\newblock \href {http://arxiv.org/abs/1703.08885} {{Question Answering from
  Unstructured Text by Retrieval and Comprehension}}.

\bibitem[{Welbl et~al.(2018)Welbl, Stenetorp, and
  Riedel}]{welbl2018constructing}
Johannes Welbl, Pontus Stenetorp, and Sebastian Riedel. 2018.
\newblock Constructing datasets for multi-hop reading comprehension across
  documents.
\newblock \emph{Transactions of the Association of Computational Linguistics},
  6:287--302.

\bibitem[{Wiese et~al.(2017)Wiese, Weissenborn, and Neves}]{wiese2017neural}
Georg Wiese, Dirk Weissenborn, and Mariana Neves. 2017.
\newblock Neural domain adaptation for biomedical question answering.
\newblock \emph{arXiv preprint arXiv:1706.03610}.

\bibitem[{Yang et~al.(2018)Yang, Qi, Zhang, Bengio, Cohen, Salakhutdinov, and
  Manning}]{yang2018hotpotqa}
Zhilin Yang, Peng Qi, Saizheng Zhang, Yoshua Bengio, William~W Cohen, Ruslan
  Salakhutdinov, and Christopher~D Manning. 2018.
\newblock Hotpotqa: A dataset for diverse, explainable multi-hop question
  answering.
\newblock \emph{arXiv preprint arXiv:1809.09600}.

\bibitem[{Yu et~al.(2018)Yu, Dohan, Luong, Zhao, Chen, Norouzi, and
  Le}]{yu2018qanet}
Adams~Wei Yu, David Dohan, Minh-Thang Luong, Rui Zhao, Kai Chen, Mohammad
  Norouzi, and Quoc~V Le. 2018.
\newblock Qanet: Combining local convolution with global self-attention for
  reading comprehension.
\newblock \emph{arXiv preprint arXiv:1804.09541}.

\bibitem[{Zhou et~al.(2018)Zhou, Cui, Zhang, Yang, Liu, and
  Sun}]{zhou2018graph}
Jie Zhou, Ganqu Cui, Zhengyan Zhang, Cheng Yang, Zhiyuan Liu, and Maosong Sun.
  2018.
\newblock Graph neural networks: A review of methods and applications.
\newblock \emph{arXiv preprint arXiv:1812.08434}.

\end{thebibliography}
\bibliographystyle{acl_natbib}

\appendix

\end{document}